\documentclass[11pt]{article}

\usepackage[margin=1in]{geometry} 
\usepackage{graphicx}
\usepackage{csquotes}
\usepackage{hyperref}
\usepackage{setspace} 
\usepackage{parskip} 
\usepackage{titlesec} 
\usepackage{booktabs} 
\usepackage{amsmath} 

\hypersetup{
    colorlinks=true,
    linkcolor=blue,
    urlcolor=blue,
    citecolor=blue
}

\title{A Pipeline for Generating Longitudinal Synthetic Clinical Notes Using Large Language Models}

\author{
    William Poulett\thanks{NHS England Data Science and Applied AI Team. Email: \texttt{england.datascience@nhs.net}}
    \and
    Alice Waterhouse\footnotemark[1]
    \and
    Ben Wallace\footnotemark[1]
    \and
    Scarlett Kynoch\footnotemark[1]
    \and
    Amaia Imaz Blanco\footnotemark[1]
    \and
    Michael Spence\footnotemark[1]
    \and
    Jonathan Pearson\footnotemark[1]
}

\begin{document}

\maketitle

\begin{abstract}

Synthetic data is increasingly used to enable the development and evaluation of AI systems in domains where access to real-world data is restricted. In healthcare, clinical documentation presents particular challenges due to its sensitivity. This work introduces a synthetic clinical notes pipeline and dataset designed to support the development of clinical AI tools while avoiding the privacy risks associated with real patient data.

The dataset is generated using a modular pipeline that combines structured patient generation, semi-structured patient journey simulation, and unstructured clinical note generation using large language models. The pipeline is designed to prioritise internal consistency across longitudinal patient records, while also capturing variation in writing style, note structure, and clinical detail. Additional mechanisms, including LLM-based validation and augmentation steps, are used to improve faithfulness, realism, and diversity of the generated notes.

We release a dataset of 70 synthetic patients, each associated with 20–50 clinical notes spanning a full hospital journey. The dataset is provided at multiple levels of validation, enabling users to balance realism and scalability depending on their use case.

This dataset supports the development, testing, and evaluation of clinical AI systems, including summarisation tools, coding models, and decision support systems, without reliance on real patient data. Our code is available on \href{https://github.com/nhsengland/synthetic_clinical_notes/tree/main}{GitHub} and our data on \href{https://huggingface.co/datasets/NHSEDataScience/synthetic_clinical_notes}{Hugging Face}.

\end{abstract}

\section{Introduction}

Clinical AI systems, including summarisation tools, diagnostic support systems, and automated coding models, require large volumes of clinical documentation for development and evaluation. Access to this data is heavily restricted. Patient records are subject to stringent regulatory and ethical requirements, and institutional approval processes for using them in research are often slow or unsuccessful. Privacy-enhancing techniques such as anonymisation and de-identification can reduce risk, but they do not always satisfy approval requirements. \cite{narayanan2008robust}

Synthetic data offers a practical alternative. Synthetic clinical notes are generated from scratch to reflect the structure and content of real clinical documentation, without containing genuine patient information. Recent advances in large language models (LLMs) have made high-quality synthetic text generation significantly more accessible, with growing interest in applications to clinical NLP  \cite{MURTAZA2023100546, PEZOULAS20242892}.

Generating synthetic clinical notes that are useful for AI development requires careful design. Notes must be medically plausible and structurally realistic. They must also be internally consistent across a longitudinal patient record, where multiple notes describe the same patient, the same events, and the same evolving clinical picture. Additionally, the dataset must reflect the range of variation present in real clinical settings, including differences in writing style, level of detail, specialty-specific terminology, and clinician documentation habits.

This paper presents a modular pipeline for generating synthetic clinical notes and a publicly available dataset produced using that pipeline. The pipeline incorporates structured patient generation, longitudinal journey simulation, and LLM-based note generation with validation and augmentation steps. We release a dataset of 70 synthetic patients, each with 20 to 50 clinical notes spanning a full hospital journey, provided at multiple levels of validation to support different downstream use cases.

Our contributions are as follows:

\begin{itemize}
    \item A modular, configurable pipeline for generating synthetic clinical notes with longitudinal consistency
    \item A publicly available dataset of synthetic clinical notes across 70 patients, provided at multiple validation levels
    \item An evaluation of pipeline outputs across dimensions of realism, consistency, and variation
\end{itemize}

\section{Dataset Description}

\subsection{Overview}

We plan to release the dataset at three levels of validation: Bronze, Silver, and Gold. Currently, only silver has been released. The Gold dataset will consist of data generated using our validated pipeline and LLM, with additional review and validation by clinicians. The Silver dataset consists of data generated using our validated pipeline and LLM, without clinician review. The Bronze dataset will consist of data generated using our validated pipeline, possibly with new LLMs, without additional validation steps.

\subsection{Known Issues}

Some special characters may be incorrectly decoded in the dataset. Users are advised to perform basic cleaning before use.

\subsection{Silver Dataset}

The Silver dataset contains data generated using our pipeline with GPT-4o as the underlying language model. It includes 70 patients, each associated with approximately 20–50 synthetic clinical notes. Of these patients, 50 are adults and 20 are paediatric cases.

\subsection{Data Structure}

The dataset is provided across three tables: 

\begin{enumerate}
    \item \texttt{patients.csv}
    \item \texttt{admissions.csv}
    \item \texttt{synthetic\_clinical\_notes.csv}
\end{enumerate}

The \texttt{patients.csv} table contains demographic and identifying information for each synthetic patient, including age, date of birth, full name, gender identity, NHS number, and a unique person identifier.

The \texttt{admissions.csv} table contains admission-level information for each patient, including admission identifiers, admission method and timestamp, admission title, bed location, patient identifiers, site information, and ward details.

The \texttt{synthetic\_clinical\_notes.csv} table contains the generated clinical notes associated with each patient admission. This includes metadata such as timestamps, note identifiers, note subject and type, as well as the clinical note text itself.





\section{Related Work}

Synthetic data can enable a range of use-cases including end-to-end software testing; tool demonstration; faster innovation; novel linkage; evaluation of solutions; and addressing bias and quality. The choice of approach greatly depends on the type of data to be synthesised, the use-case, and the balance between privacy, fidelity, fairness, explainability and adoption.  

To ensure adequate coherence across different notes in the patient journey, we needed a methodology which clearly defines the patient, their reason for being in hospital, and the series of events that make up their journey. There are therefore three different types of synthetic data which we needed to consider:

\begin{enumerate}
    \item The patient demographics and their reason for hospitalisation can be represented by structured data;
    \item The patient’s journey through hospital is semi-structured;
    \item The final outputs, i.e. the clinical notes themselves, constitute unstructured data.
\end{enumerate}

\subsection{High Privacy Structured Data Generation}

Synthetic tabular data can either be generated from raw data or domain knowledge. If generating from raw data, then the first step is to create a representation which captures the content and variability of the raw data but also balances this against privacy so no individual data point can be confidently identified from the representation. The second step is to then sample from this representation in a way that considers inherent biases.

The simplest way of doing this is to “erode” raw data (through adding noise, suppressing low counts, reallocation and aggregation) until a privacy threshold (usually defined by k-anonymity, l-diversity, or t-closeness \cite{DBLP:journals/corr/abs-2104-06523} is achieved. The NHS Digital Artificial Data pilot is one such example of this \cite{nhs_artificial_data}.  

Alternatively, we can use tools such as Synthea \cite{synthea_mitre}. This is an open-source patient population simulator which allows for the generation of machine readable, realistic and synthetic patient data. This can then generate lifespans of realistic but fictional synthetic patient electronic health records. Alternative simulation engines including agent-based modelling (Agent Hospital \cite{li2025agenthospitalsimulacrumhospital}) and digital twin setups which have been developed but often turn out to be too burdensome to maintain for general usage.

The evaluation of the tabular synthetic data generated includes fidelity, utility, fairness and privacy metrics.

\begin{itemize}
    \item To measure fidelity, we turn to statistical tests to understand profile comparisons (Pearson’s or Spearman’s correlations), distribution comparisons (Kolmogorov-Smirnov test, Chi-Squared tests), detection metrics (logistic regression or Support vector machines), Variance metrics (e.g. Voas-Williamson or propensity scores), Aggregate difference metrics (e.g. KL divergence, Gower distance, or Wasserstein metric), and off-manifold and latent space checks (PCS and T-SNE). 
    \item Utility is best measured through application to a downstream task with a clear definition of accuracy.  
    \item Fairness can be measured and addressed through pre or post processing but ideally as a constraint within the model forcing the generation to adhere to equalised odds or demographic parity.  
    \item As mentioned previously, privacy is often measured through conducting k-anonymity or similar metrics after the generation of the data or through the use of differential privacy within the model itself.
\end{itemize}
To support the evaluation of synthetic data, frameworks such as SDV \cite{datacebo_sdv_dev} have been developed with a range of these metrics and methods built in.

\subsection{High Fidelity Semi-Structured Data Generation with Domain Knowledge}

Assuming the patient demographics are generated with high fidelity from the previous step, we now turn to creating a realistic patient journey for each individual. One approach is rule-based template extraction: simulation engines like Synthea inherently produce time-stamped events (e.g. “patient admitted with pneumonia, received antibiotic X on day 3”). A Synthea UK \cite{nhs_swpc_synthea} version also exists but is focused on primary care.

LLMs are general-purpose models capable of performing a wide range of language-based tasks. Whilst traditionally strongest in unstructured data generation, they can increasingly be used for semi-structured and even structured data generation. One example is within knowledge-graph–guided sampling: for example, the MedSyn \cite{Kumichev_2024} framework builds a Medical Knowledge Graph (from sources like UMLS or ontologies) and samples related facts (symptoms, findings, comorbidities) from it to include in the LLM prompt for generation.

Another useful example is that LLMs can simply be prompted to give structured JSON outputs. However, LLM structured outputs often fail to compile and need their outputs cleaning to be useful. Recently, many LLM providers included the ability to force structured outputs in LLM responses which has reduced the risk of compilation failure drastically.

\subsection{Highly Adaptable Unstructured Data Generation with LLMs}

LLMs are increasingly being applied to medical domains. Their effectiveness in medical applications - particularly in generating synthetic data - can be influenced by two key factors:

\begin{enumerate}
    \item The inherent performance of the LLM on medical tasks (i.e., its medical knowledge and reasoning capabilities).
    \item The prompting techniques used to guide the model's responses.
\end{enumerate}

A common error experienced by LLMs is hallucinations – generating plausible but factually incorrect content \cite{cossio2025comprehensivetaxonomyhallucinationslarge} . For many medical tasks this is problematic, particularly when using LLMs to generate discharge summaries. Various studies have been done on LLM performance on medical tasks. For example, Kim et al \cite{Kim2025.02.28.25323115} investigated the hallucination rates of several state-of-the-art LLMs and found significant disparities in performance across models. Here, OpenAI’s GPT-4o exhibited the highest hallucination rate. Some models are specifically trained for medical tasks. Google has released Med-Gemini \cite{saab2024capabilitiesgeminimodelsmedicine}, and other studies have developed domain-specific medical LLMs, such as GatorTronGPT and Me-LLaMA \cite{Peng2023}.

However, when generating synthetic data some hallucinations could be appropriate. Whilst we want all synthetic clinical data to represent plausible patient journeys and clinical documents, hallucinations may help to generate real world noise and could produce appropriate outputs useful for testing and evaluation.

There are further steps we can take to improve the quality of our clinical notes. One common method is prompt engineering: the iterative refinement of prompts to elicit better outputs. Chain-of-Thought prompting \cite{wei2023chainofthoughtpromptingelicitsreasoning}, which encourages models to reason step-by-step before answering, has been shown to also improve performance across tasks.

LLM validators can also be incorporated into LLM pipelines. These are LLMs prompted to judge and improve outputs based on specific criteria. Using validators, Gosmar et al demonstrated a successful three-agent system that detected and removed hallucinations from LLM-generated outputs \cite{Gosmar2025HallucinationMU}.

LLM performance can also be enhanced through access to additional knowledge. We define ‘additional knowledge’ broadly, ranging from in-context learning to the use of knowledge graphs. In-context learning enables LLMs to generate responses that mirror examples provided within the prompt \cite{HE2025102963}.  Retrieval-Augmented Generation (RAG) allows LLMs to search databases and retrieve relevant information, and has been shown to improve performance in medical tasks \cite{xiong2024benchmarkingretrievalaugmentedgenerationmedicine}. Some studies have further improved retrieval methods by incorporating structured resources such as knowledge graphs \cite{jiang2025reasoningenhancedhealthcarepredictionsknowledge}. Notably, many of these techniques can be used in combination for greater effect.

Alongside synthetic data generation, LLMs can also be used to evaluate the outputs of other LLMs - known as LLM-as-a-Judge \cite{gu2025surveyllmasajudge}. We have already discussed one type of LLM Judge: a validator, but LLM Judges can also grade, evaluate, critique, and rank responses . However, an LLM-as-a-Judge is not perfect. LLM Judges can show bias, including favouring the first response when comparing two outputs, or preferring longer responses \cite{ye2024justiceprejudicequantifyingbiases}. LLM judges can themselves be evaluated. One approach is to assess their alignment with human judgments using metrics such as Cohen’s kappa \cite{jiang2025reasoningenhancedhealthcarepredictionsknowledge}. 

\section{Methodology}

\begin{figure*}
    \centering
    \includegraphics[width=\textwidth]{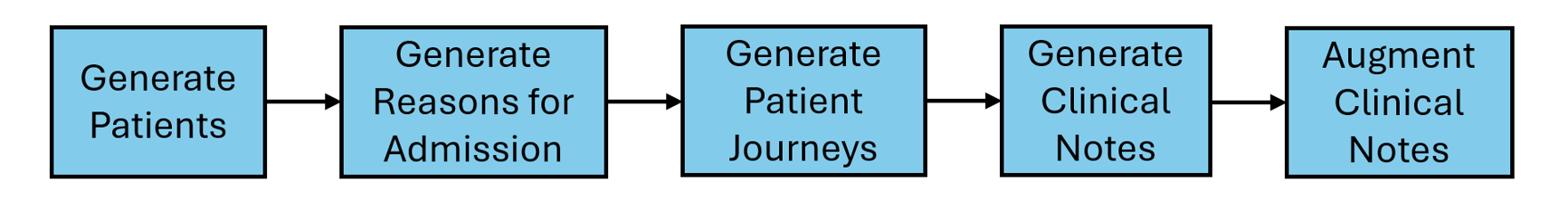}
    \caption{A simple overview of our synthetic data pipeline.}
    \label{diagram_1}
\end{figure*}

For the task in hand the main aim was to quickly create data. This therefore needed to contain realistic form and content, but the fidelity did not need to be as high as many of the aforementioned methods aim for. Instead, the privacy and timeline elements were key. We opted for an LLM-heavy approach for both the semi-structured and unstructured stages, to facilitate speed of development. This was supplemented with highly private structured data from Synthea, as well as hand-picked admission reasons, to maintain a high degree of control over the test cases produced. Information from these structured data sources was injected into our prompts in a manner comparable to the MedSyn work.

Figure \ref{diagram_1} shows an overview of the methodology we used to create the synthetic notes. There are five high-level stages to this. Alongside this pipeline is an evaluation suite to aide iterative improvement of the pipeline.

\begin{itemize}
    \item Generate Patients
    \item Generate Patient Admissions
    \item Generate Patient Journeys
    \item Generate Clinical Notes
    \item Augment Clinical Notes
\end{itemize}

\subsection{Generating Patients}

\begin{figure*}
    \centering
    \includegraphics[width=\textwidth]{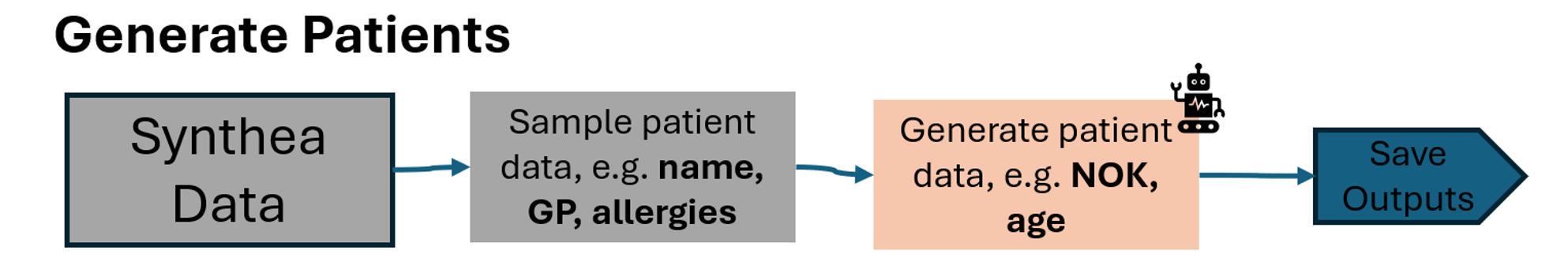}
    \caption{Generating a population of patients. Synthea data is used alongside an LLM to create a population of varied but realistic patients.
}
    \label{diagram_2}
\end{figure*}

Figure \ref{diagram_2} shows our patient generation step. In order to generate basic patient information such as name, age and gender, we use Synthea. As Synthea was originally developed in a US context, for this project we use Synthea UK , a modified version of Synthea developed by NHSE. This version is designed so that the demographics and addresses of generated patients align with the UK population.

It should be noted that Synthea is a relatively complex tool and that we only use the first step of its data generation pipeline, initial patient generation, in this work. Synthea UK is focused on primary care - to use it later in our pipeline we would require secondary care journeys and development of a secondary care simulation engine was not possible in the time available. Patient names are generated using a random combination of the most common given names and surnames in the UK.

After sampling data from Synthea, we ask an LLM (GPT-4o) to generate additional patient details using the patient prompt found in our prompt appendix. The details generated in this stage include the name and contact details of the patient’s next of kin. GPT-4o is typically used throughout the pipeline, but our pipeline is built to be reusable with other LLMs.

Additionally, all of our prompts have been iteratively refined following clinical feedback and manual review. One example of this iterative refinement is that GPT-4o would almost always assign patients an allergy to penicillin. To rectify this, we now randomly sample from a group of allergies with their national prevalence and feed this value into our LLM prompt using an allergies variable.

Throughout this pipeline, all LLM outputs are expected to be JSONs, or lists of JSONs. Whilst LLMs are typically quite good at generating outputs in the specified format, our pipeline utilises a function to clean outputs when the LLM output is not a valid JSON. This function first attempts to extract the correct JSON output using Regular Expressions (regex). If this fails, an LLM is prompted to clean the output. It is given the output alongside the corresponding error which occurs when attempting to decode the output using the python JSON library.

\subsection{Generating Admissions}

\begin{figure*}
    \centering
    \includegraphics[width=\textwidth]{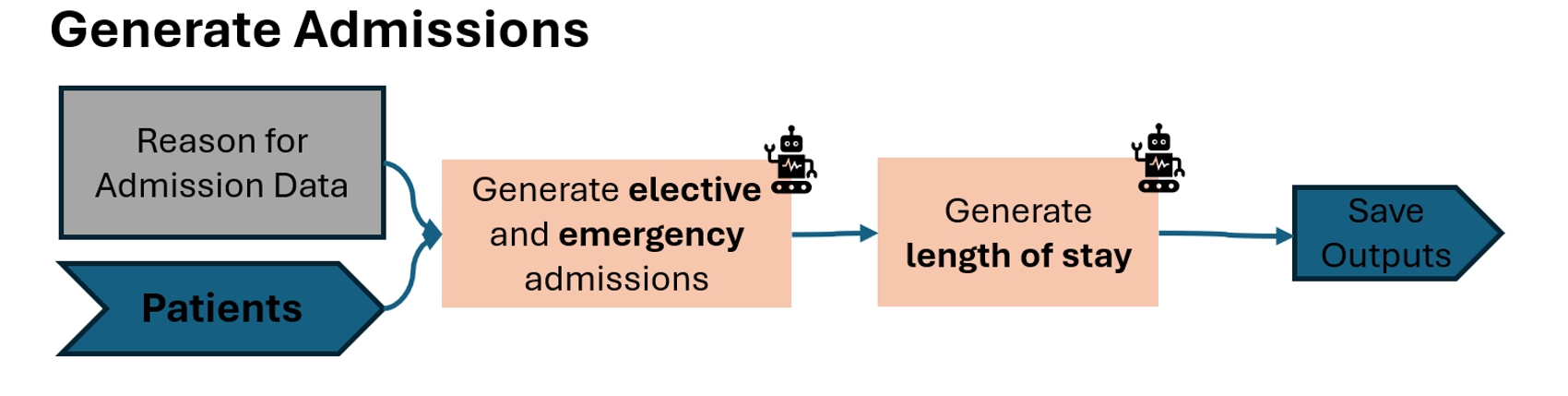}
    \caption{Generating reasons for admission for patients. Admissions can either be elective admissions or emergency admissions.}
    \label{diagram_3}
\end{figure*}

As seen in \ref{diagram_3}, our pipeline can generate two types of admission: elective and emergency. Each admission uses a slightly different prompt that has been iteratively improved based on clinical feedback. The admission reason is sampled from a table containing common admission reasons for different sexes and age bands. For our use case, this table was a hand-picked list of admission reasons for which test cases were required. However, the table could easily replaced with an aggregate dataset of SNOMED or ICD coded reasons for admission to provide greater variability in the synthetic data outputs.

The admission details that are generated by an LLM include the specialty and ward the patient is admitted to, as well as their current medications and past medical history. Once the patient admission details have been generated, another LLM call is used to estimate the patient length of stay.

\subsection{Generating Patient Journeys using an LLM}

\begin{figure*}
    \centering
    \includegraphics[width=\textwidth]{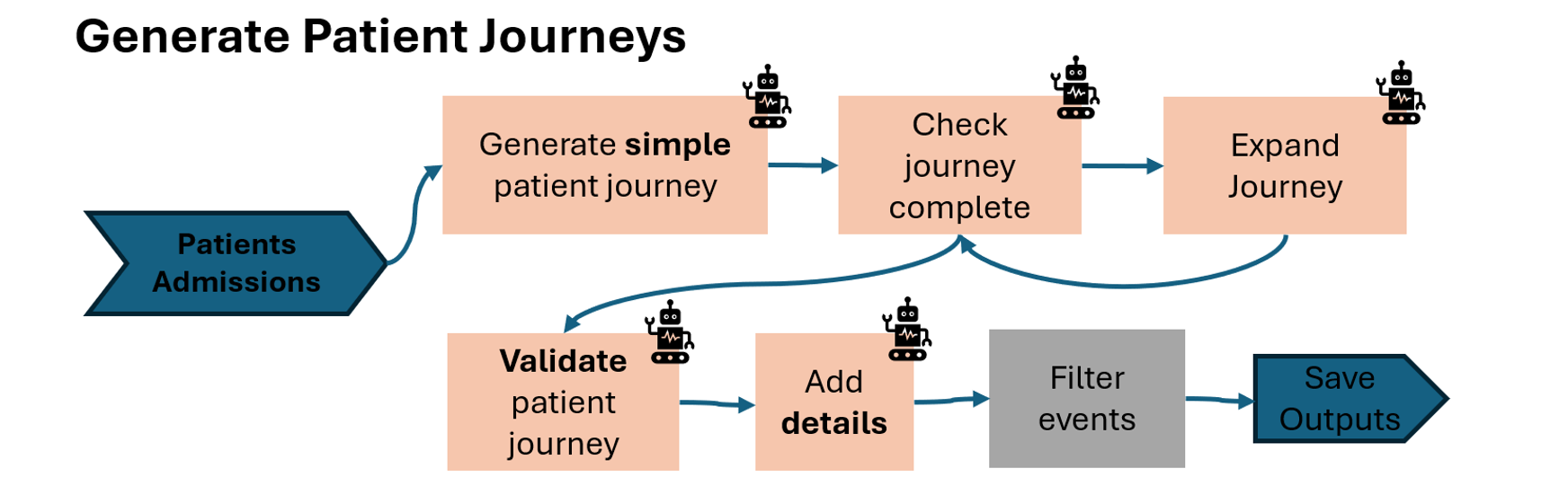}
    \caption{Generating patient journeys. Patient journeys are generated using a series of LLM calls.}
    \label{diagram_4}
\end{figure*}

This stage (\ref{diagram_4}) of the pipeline uses multiple LLM calls to generate realistic patient journeys. A patient journey is a set of events starting from the point of admission up until the moment just before discharge. This method was chosen due to its adaptability and no reliance on real data.

First, a ‘simple’ journey is generated using an LLM prompt which includes the reason for admission, patient information and estimated length of stay. A ‘simple’ journey only includes the event type, date, time and a short summary. We later add complexity to each event to provide better grounding for the clinical note, however, adding all the complexity in one LLM call was found to decrease the quality of the patient journeys.

The event types we ask the LLM to use when generating the first simple journey include:

\begin{enumerate}
    \item Admission
\item Post-take ward round (the first ward round after admission)
\item General ward round
\item Operation
\item Registrar review
\item Consultant review
\end{enumerate}
  
We found that for longer journeys, the LLM often cuts-off before completing the full journey.

\begin{displayquote}
    \textit{For brevity I have stopped generating events. Would you like me to continue?}
\end{displayquote}

To deal with this, another LLM prompt detects journeys that are terminated early. If found, the LLM is asked to continue the journey until a full simple journey is complete.

Once a simple journey is complete, an LLM validator assesses the realism of the journey. If unrealistic, it has the power to rewrite certain sections of the journey to increase realism. This section of the pipeline can loop multiple times.

Next, complexity is added to each event. This includes adding staff members to each event, more detailed descriptions and next-step decisions.

Occasionally, events within the journey include events we are not interested in. A common example is the discharge itself. As we do not want to generate our own discharge summary, we filter out unwanted events.

The prompts in this section have been carefully designed following clinician feedback. Like other sections of the pipeline, LLM outputs are regularly cleaned to ensure they output valid JSON.  

\subsection{Generating Clinical Notes using LLMs}

\begin{figure*}
    \centering
    \includegraphics[width=\textwidth]{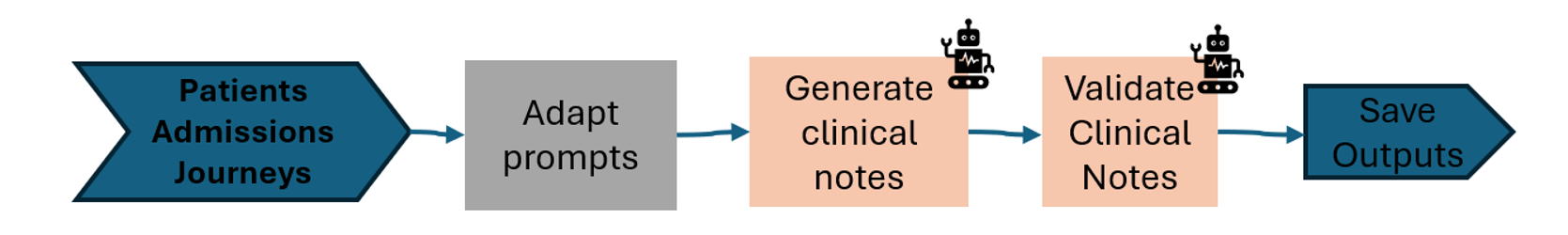}
    \caption{Generating clinical notes. Our clinical note generation stage uses very versatile prompts to tailor the prompt to each note type, patient and event. Once clinical notes are generated, a validator checks the faithfulness of the note to the event description.}
    \label{diagram_5}
\end{figure*}

This stage (\ref{diagram_5}) of the pipeline uses the detailed description of each event to generate a single clinical note per event.

\subsubsection{
Generating Clinical Notes from Patient Journeys
}

For each patient, each clinical note is generated by prompting GPT-4o with the patient information, a list of patient events to have occurred up until the current event, and an expected output format. The output schema for each clinical note type we generate was designed with help from a clinician to ensure realism.

The prompt is very versatile and has many different variations depending on the event type. For example, we learnt from clinicians that in patient handovers and during admission, patients are checked for ‘red-flag’ symptoms. The symptoms that should be checked depend on the admission reason, and we therefore use an LLM to select which red flags are appropriate. When red flags should be checked, an extra instruction is included in the prompt to describe the process of checking these symptoms in the clinical note. Similarly, another LLM prompt helps select the relevant patient examinations required by law to be done during examination events.

In addition, clinical notes are written in different styles depending on the member of staff writing the note.

\subsubsection{Clinical Personas}

Following feedback from clinical reviewers, we added to our pipeline the option of incorporating clinical personas to increase the real-world noise found in clinical notes. After the generation of a patient journey, each member of staff is assigned a random ‘persona’. These are:

\begin{enumerate}

\item \textbf{Concise}: Use very short, clear sentences. Avoid filler words and redundancy. Ensure all details are written down but there is no extra information. Ensure each section is as short as possible
\item \textbf{Narrative}: Write in a flowing, story-like manner. Connect ideas with transitions and natural pacing
\item \textbf{Bullet Points}: Present information as bullet points for easy readability. Bullet points are written using '-', but write nothing else in markdown form. You do not need to use full sentences, each bullet point can be in concise note form.
\item \textbf{Note}: Write in incredibly short notes. Notes need to only make sense to fellow medical staff, so can be include abbreviations and shortened words. Ensure each section is short. You may leave sections blank if they are not necessary.
\item \textbf{abcde}: Write in short and concise note form. When describing patient physical examinations, write notes using the ABCDE approach. Create 5 short bullet points - one for each section labelled: A,B,C,D,E. Ensure each section is short.
\end{enumerate}

Originally, we included other personas such as:
\begin{enumerate}
\item \textbf{Verbose}: Be detailed and expansive. Explain ideas thoroughly, even if it takes more space.
\end{enumerate}

When each clinical note is generated, the pipeline will identify the first member of staff at that event and write the clinical note with their persona included in the prompt. If the staff member was not in the original prompt and has no persona, a new one is generated. This means a staff members notes will have a consistent style throughout a patient journey

\subsubsection{Using LLM Validators}

After the note is generated, we use an LLM validator to detect how faithful a note is in the patient information and patient journey and make changes to the clinical note if deemed necessary. This validator is built into the pipeline within a loop, meaning a validator can validate the previous validators output to continually improve the output. Research has found that multiple AI reviewers are better than single AI reviewers at detecting hallucinations.

The validator prioritises faithfulness: how accurately the note includes all vital information from the provided event and patient context, ensuring no critical details are omitted or altered. As our clinical notes are longer than the patient journey summaries they’re based on, we expect some additional clinically realistic detail. This is acceptable, provided it is consistent with the event. What matters most is that every detail in the event is accurately included in the note.

Variables in the validation prompt are the clinical document, the patient information, and the full patient journey up to that event. 

\subsection{Augmenting Clinical Notes}

\begin{figure*}
    \centering
    \includegraphics[width=\textwidth]{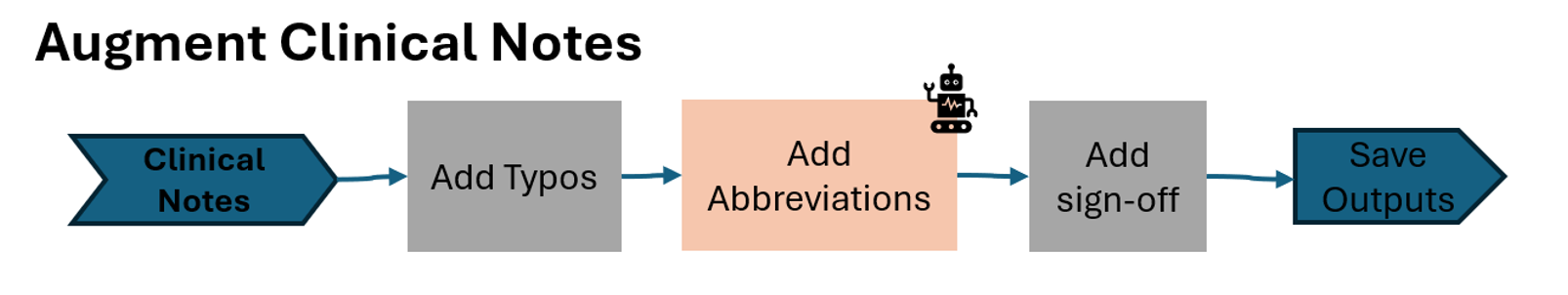}
    \caption{We augment our clinical notes by optionally adding typos, abbreviations, and staff signoffs.}
    \label{diagram_6}
\end{figure*}

As seen in \ref{diagram_6}, this stage of the pipeline adds abbreviations, adds typos and adds sign-offs to notes.

\subsubsection{Adding Abbreviations}

Originally, a list of common clinical abbreviations alongside their meanings was compiled by a clinician. A simple find and replace function was used to replace phrases from this list with a corresponding abbreviation according to a replacement probability. 

However, we found that some abbreviations have meanings too complex to simply use a find and replace for. For that reason, we designed an LLM prompt to add abbreviations to a string. As it is an LLM, it can add abbreviations that flow in the sentence.

\subsubsection{Typo Generation}

Following clinical feedback, we used the python typo package \cite{typo_pypi} to add realistic typos to our clinical notes. This package aims to simulate realistic typos by using techniques such as:

\begin{enumerate}
    \item Swapping two consecutive characters.
    \item Replacing characters with a character close to them on a keyboard.
    \item Randomly removing characters.
    \item Adding double spaces.
\end{enumerate}

Each member of staff is assigned a random typo rate, which means some staff will consistently have more typos throughout a patient journey in comparison to others.

\section{Evaluation and Iterative Improvement}

\begin{figure*}
    \centering
    \includegraphics[width=\textwidth]{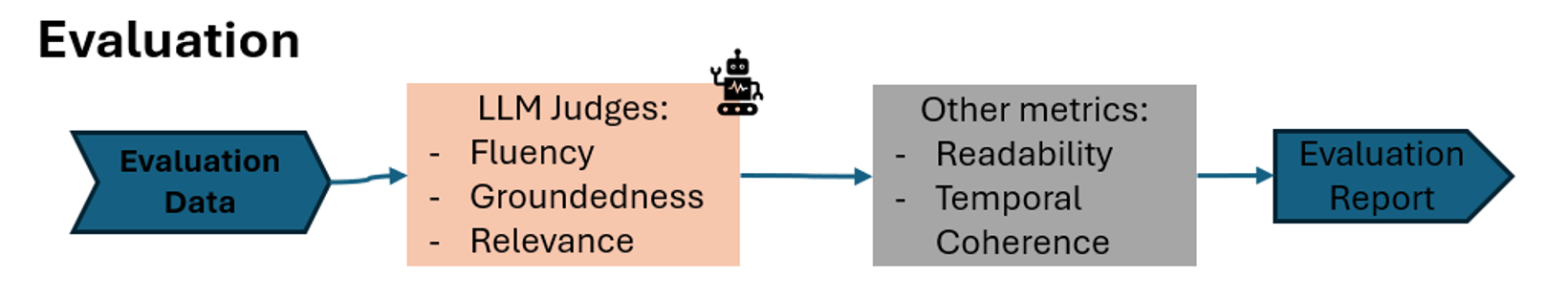}
    \caption{Generation of an evaluation report using traditional metrics and LLM Judges.}
    \label{diagram_7}
\end{figure*}

We built an evaluation pipeline (\ref{diagram_7}) to evaluate the quality of our clinical notes during development, allowing us to iteratively improve our prompts and pipeline for better clinical notes.

\subsection{Readability}

The first metric we measure is the readability of the clinical notes. This is important as we expect to develop clinical personas in the future, where different personas write clinical notes in different styles. Some notes will be in short form bullet points, while others long prose. We expect these different personas to have differing levels of readability, so want to measure the effect of this over time with a readability score.

We have selected the Flesch Reading Ease Score and Dale Chall Readability Score from textstat for our Evaluations.
The Flesch Reading Ease Score considers the total words, total sentences and total syllables to measure the readability. Number range from 0 to 100, with 0 being difficult to read and 100 being very easy to read.

The Dale Chall Readability Score uses a lookup table of the most commonly used English words. Scores of 4.9 or lower are easier to read (approximately a 4th-grade student), and 9+ is hard to read. As we have used textstat, it is very easy to substitute in more readability formula.
To summarise the score for the entire dataset, we take the mean average of both readability formula.

\subsection{LLM-as-a-Judge}

We have used an LLM to score the fluency, groundedness and relevancy of each clinical document.

\begin{enumerate}
    \item  Fluency refers to how naturally and smoothly the text reads, considering factors such as grammar, coherence, sentence structure, and ease of understanding.
    \item Groundedness refers to how well the note aligns with the given event and patient information, ensuring that it is factually consistent and logically supported.
    \item Relevance refers to how well the note captures and includes all vital information from the provided event and patient information, ensuring no critical details are omitted.
\end{enumerate}

We measure the fluency as another method to ensure the effectiveness of our clinical personas. We measure groundedness to ensure clinical notes are true to the patient journey and patient information we have generated. Finally, we measure relevance to ensure no critical information is missed when generating a clinical note.

Each Judge scores the metric on a scale of 1 to 5, with 1 being a poor performance on the metric and 5 being a perfect performance. To summarise the score for the entire dataset, we take the mean score from each LLM Judge across all clinical notes.

\subsection{Temporal Coherence}

We want to ensure each clinical note is dated such that events occur in order. This is a simple test that looks at the time stamps of each event for each patient, and checks whether the next event occurs at the same time or after.
Each patient returns a True if all events are in order, or a False if any events are not. To calculate a score for the entire dataset, the fraction of patients where all events are in time order is returned.

\subsection{Human Evaluation}

To increase trust in testing that was done with the synthetic data, we got several rounds of feedback from human clinicians to evaluate how realistic the clinical notes were. This identified a number of issues, which we fixed by making changes to our pipeline and prompts. Some examples of this are shown in the table below.

\begin{table}[h]
\centering
\begin{tabular}{|p{0.4\textwidth}|p{0.6\textwidth}|}
\hline
\textbf{Issue Identified} & \textbf{Improvement Implemented} \\
\hline
Some note types that are present in real-world documentation were missing & Add additional event types to our journey generation, with corresponding output templates for clinical notes \\
\hline
Red flag symptoms were not documented as having been ruled out & Add an additional LLM call to identify red flag symptoms and pass these into the note generation prompt with instructions to ensure that they are ruled out \\
\hline
Some reasons for admission were not adequately specified in the notes, e.g. sepsis with no identified cause & Ensure that the list of reasons for admission included enough detail to fully specify the problem \\
\hline
Notes were too verbose & Prompt engineering \\
\hline
Note structure and subheadings were unrealistic & Modifying the expected output template which is passed into the LLM prompt \\
\hline
It was unclear what the role of the clinicians writing the notes was & Be clearer about the roles of synthetic clinicians and include sign-offs in the templates for each note type \\
\hline
Some abbreviations were unrealistic & Change the approach to abbreviations to use an LLM \\
\hline
\end{tabular}
\caption{Identified issues and corresponding improvements}
\label{tab:improvements}
\end{table}

\section{Additional Capabilities}

There are additional capabilities built into our pipeline which can be used when generating data.

\subsection{Bias Testing}

This project contains a bias testing notebook which follows a methodology proposed by Rickman \cite{Rickman2025} to swap the gender of synthetically generated clinical notes.  Evidence demonstrates that LLMs can produce biased outputs in summarisation tasks, particularly with respect to protected characteristics such as gender.

LLM's are used to change the gender of the patient when referred to in clinical notes. This allows LLM Judges to detect variation between discharge summaries written about male and female patients - possibly identifying bias.

\subsection{Novel and Rare Disease Sampling}

Our pipeline also has the ability to sample rare and novel diseases. These can be used to test edge cases with rare scenarios.

\section{Limitations and Future Work}

There were several possible improvements to the pipeline that we identified through clinician feedback and internal review, but did not prioritise due to time constraints. These included:

\begin{enumerate}
    \item Adding an option to make a certain percentage of journeys “complex”, so that e.g. the initial clinician impression is different to the final diagnosis, or the patient experiences a treatment complication such as a hospital-acquired infection
    \item Tracking which clinician persona was used to write the note, to aid in evaluating outputs
    \item Including a configurable parameter to control whether information about previous events is included in the note prompt and note validation prompt
    - Changing the ward round templates throughout the patient stay so that later ward round notes are more concise
    \item Experimenting with different amounts of abbreviations for different personas
    \item Improving the way discharge planning considerations such as frailty and community referrals are documented
    \item Adding an event type for minor procedures like lumbar punctures, which have different documentation requirements to operations
    \item Reducing the frequency of nursing and therapies events to be more realistic
    \item Improving the templates for nursing and therapy notes to follow a SOAP format (Subjective, Objective, Assessment, Plan)
\end{enumerate}

\section{Ethical Considerations}

We would like to make clear that all data in this pipeline is entirely synthetic. Synthetic data is artificially generated data that replicates the statistical properties or qualities of real-world datasets.

Typically, it is generated using real data and adding noise. However, in this pipeline \textbf{no real data is used} at any point.

Synthetic data is incredibly useful; it offers advantages in privacy preservation, scalability, and cost reduction. However, there are various limitations that are important to understand, particularly for our use case of generating synthetic clinical notes:

\subsection{Synthetic Data Limitations}

\begin{enumerate}
    \item \textbf{Loss of Real-World Complexity}
    Real-world clinical data is complex. A patient journey through hospital can be influenced by endless factors - from conflicting observations to staff availability. Meanwhile, clinical documentation is shaped by the implicit knowledge of staff, institutional practices and context-dependent decision-making.
    
    Whilst our pipeline can generate high quality synthetic journeys and notes, it will fail to capture this richness.
    \item \textbf{Underrepresentation of Edge Cases}
    Rare scenarios can include atypical admissions, conflicting diagnoses, or adverse events which are difficult to generate reliably.

    Whilst our pipeline intends to be able to generate a variety of synthetic clinical journeys and notes, these may tend to reflect 'average' cases rather than the long tail of real-world complexity.
    \item \textbf{Bias Amplification}
    Synthetic clinical notes can inherit biases from both the underlying Large Language Model (LLM) and the prompting strategy. These biases can manifest in:
    \begin{itemize}
        \item Overemphasis on certain conditions or treatments
        \item Systematic omission of specific patient groups or outcomes
        \item Stylised or homogenised documentation patterns
    \end{itemize}
    \item \textbf{Hallucinations}
    LLMs generating synthetic notes and journeys may introduce plausible but incorrect clinical information. These hallucinations can include:
    \begin{itemize}
        \item Fabricated symptoms or test results
        \item Inaccurate timelines
        \item Inconsistent patient states across notes
    \end{itemize}
\end{enumerate}

\section{Conclusion}

This document describes the methodology behind our synthetic clinical note and dataset generation. The process follows the latest research in LLM usage and evaluation and has been iteratively refined following clinical feedback of the clinical notes.

We believe there are various use cases for this pipeline. These include creating synthetic notes to test:

\begin{enumerate}
    \item AI generated discharge summary tools
    \item Clinical coding tools
    \item Sentiment analysis tools
    \item Safeguarding tools
    \item Multimodal predictive models such as length of stay prediction
\end{enumerate}

In addition, parts of the pipeline could be reused e.g. to generate synthetic patient journeys for other use cases, or to add noise to existing text data for testing or other purposes.


\section{Acknowledgements}

This work was completed by multiple team members in NHS England's Data Science and Applied AI Team. Thank you to all involved.

\bibliographystyle{unsrt}
\bibliography{references}

\end{document}